\documentclass{article}
\usepackage{iclr2027_conference,times}

\usepackage[hyphens]{url}
\usepackage{hyperref}
\hypersetup{hidelinks}
\usepackage{graphicx}
\usepackage{amsmath}
\usepackage{booktabs}
\usepackage{tabularx}
\usepackage{colortbl}
\usepackage{multirow}
\usepackage{caption}
\usepackage{placeins}
\usepackage{needspace}
\usepackage{wrapfig}
\usepackage{tcolorbox}
\tcbuselibrary{skins}
\usepackage{booktabs}
\usepackage{array}
\usepackage{tabularx}
\usepackage{dcolumn}
\usepackage{fix-cm}
\usepackage{xcolor}
\usepackage{colortbl}
\usepackage{threeparttable}

\definecolor{RSTBlue}{HTML}{7299D5}
\definecolor{RSTGreen}{HTML}{85BB5D}
\definecolor{RSTHeader}{HTML}{ECF1FB}
\colorlet{RSTOurs}{blue!8}
\definecolor{RSTLine}{HTML}{555555}
\definecolor{RSTMuted}{HTML}{5C646C}

\newcommand{\RSTna}{\multicolumn{1}{c}{-}}
\newcommand{\RSThead}[1]{\multicolumn{1}{c}{\textbf{#1}}}

\newcommand{\RSTNumberFont}{\small\rmfamily\upshape}
\newcommand{\gain}[1]{\textcolor{red!85!black}{\textbf{#1}}}

\newcommand{\RSTTableSetup}{%
  \small\rmfamily\upshape
  \renewcommand{\arraystretch}{1.14}%
  \setlength{\tabcolsep}{4pt}%
  \setlength{\heavyrulewidth}{0.7pt}%
  \setlength{\lightrulewidth}{0.4pt}%
  \setlength{\cmidrulewidth}{0.35pt}%
  \arrayrulecolor{RSTLine}%
}

\newcommand{\RSTciteBlink}{\citep{fu2024blink}}
\newcommand{\RSTciteMuir}{\citep{wang2024muirbench}}
\newcommand{\RSTciteVlm}{\citep{zhang2025vlm2bench}}
\newcommand{\RSTciteQwen}{\citep{bai2025qwen25vl}}
\newcommand{\RSTciteEureka}{\citep{meng2025mmeureka}}
\newcommand{\RSTcitePerl}{\citep{perl2025}}
\newcommand{\RSTciteMico}{\citep{mico2025}}
\newcommand{\RSTciteCgc}{\citep{cgc2026}}

\makeatletter
\def\fps@figure{!t}
\def\fps@table{!t}
\makeatother
\newcommand{\rev}[1]{#1}

\title{ReSynth: Constructing Synthetic Supervision for Multi-Image Understanding}

\author{%
\normalfont
Yilin Wang\textsuperscript{1}\quad
Xin Yu\textsuperscript{2}\quad
Xinyi Guo\textsuperscript{3}\quad
Xiangxi Zheng\textsuperscript{4}\\
Dongxing Mao\textsuperscript{3}\quad
Linjie Li\textsuperscript{5}\quad
Zhengyuan Yang\textsuperscript{6}\quad
Alex Jinpeng Wang\textsuperscript{3}\\[0.5em]
{\normalfont\textsuperscript{1}ZJU\quad
\textsuperscript{2}Columbia\quad
\textsuperscript{3}CSU\quad
\textsuperscript{4}NJU\quad
\textsuperscript{5}UW\quad
\textsuperscript{6}Microsoft}%
}

\iclrfinalcopy

\begin{document}

\maketitle
\lhead{}

\begin{abstract}
People often understand the world by comparing different views of it. Vision-language models (VLMs), however, receive these views as separate images and still struggle to relate them.
One problem is data. Training on these tasks requires groups of images with a known relationship: the same object from different viewpoints, or the same scene before and after a change. Such examples are hard to collect at scale from existing photo collections. We introduce ReSynth, which \rev{turns single-image perception and controllable image generation into cross-image relational supervision}. Localized editing produces before-and-after pairs, viewpoint generation produces multiple views of the same instance, and verified scene contents support comparisons of counts and attributes.
ReSynth \rev{combines Qwen3-VL-8B's captioning, grounding, and counting capabilities with Qwen-Image-2.1's generation and editing capabilities}. It verifies \rev{single-image facts such as object counts and attributes, then combines these facts to construct multi-image QA}.
Using GRPO on approximately \rev{440K} ReSynth questions, with no real images during training, Qwen2.5-VL-7B \rev{and Qwen3-VL-8B} improve by \rev{\textbf{17.73} and \textbf{6.10} average} points on BLINK, MUIRBench, and VLM2-Bench, while single-image performance \rev{is broadly preserved on Qwen2.5-VL-7B and improves across all four evaluated benchmarks on Qwen3-VL-8B}. \rev{Notably, this supervision enables the 8B perception model itself to outperform Qwen3-VL-235B on BLINK and MUIRBench.}
\end{abstract}

\section{Introduction}

Vision-language models (VLMs) have become increasingly capable at understanding individual images.
In the real world, however, visual understanding often requires comparing observations across viewpoints and time.
We may recognize the same suitcase from a new angle, spot what changed between two photos of a desk, or match a detail in one image to its counterpart in another.
Such judgments require linking evidence across images and remain challenging for VLMs on BLINK~\citep{fu2024blink}, MUIRBench~\citep{wang2024muirbench}, and VLM2-Bench~\citep{zhang2025vlm2bench}.

\begin{figure}[!t]
    \centering
    \includegraphics[width=\linewidth]{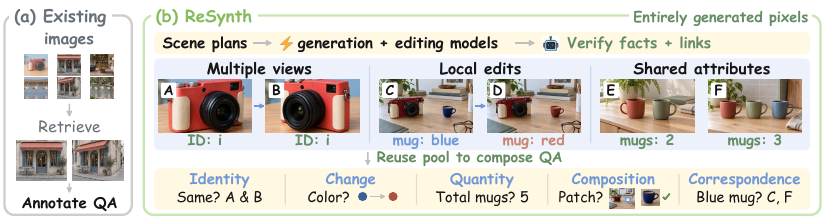}
    \caption{Existing-image curation retrieves and annotates available images.
    ReSynth plans and generates candidate relations, verifies the resulting visual evidence, and reuses it to construct questions across seven multi-image abilities; the diagram illustrates five of these task groups.}
    \label{fig:intro_comparison}
\end{figure}

Training models for these tasks requires image groups with known relationships, such as multiple views of one instance or a scene before and after a local change.
Such groups are difficult to collect at scale from existing photographs.
Multi-image instruction datasets commonly assemble or retrieve available images before constructing QA~\citep{jiang2024mantis,li2025smir,li2024llavainterleave}.
Other methods organize those images into permutations, contrastive groups, or grounded compositions~\citep{perl2025,mico2025,cgc2026}.
These approaches remain limited by the relationships represented in their source images: rearranging images cannot recover a viewpoint or change that was never captured.
Figure~\ref{fig:intro_comparison} contrasts this dependence on existing images with ReSynth's generation-based approach.

Existing photographs provide only the relationships that happened to be captured, whereas generative models allow us to construct candidate cross-image comparisons deliberately.
We can plan source images with different objects, settings, and physical relations, then use local editing to create before-and-after states or viewpoint generation to show the same subject from another angle.
\rev{While cross-image reasoning remains challenging, VLMs already have strong capabilities in single-image captioning, grounding, and counting, and image generators can synthesize controlled changes and views.}
Yet specifying a relation does not guarantee that it appears in the rendered images, and related images do not by themselves provide questions with well-supported answers.
This raises two linked questions: \textbf{can \rev{we turn single-image perception and image generation into} multi-image supervision, and how can relations visible in those images be verified and reused to construct reliable QA?}

\Needspace{4\baselineskip}
Turning generated images into supervision requires broad content coverage and consistency between planned relations, rendered images, and QA answers.
This raises three challenges.
First, unconstrained caption generation with a large language model (LLM) can concentrate on a small set of recurring objects and scenes, limiting the content coverage of source images.
Second, generation intent does not always match the resulting images: source images may fail to exhibit the specified attributes or counts, local edits may alter unrelated content, and new viewpoints may fail to preserve subject identity.
Visible image attributes and cross-image associations therefore require verification to provide a reliable basis for subsequent QA construction.
Third, reusable evidence does not by itself specify a training task; suitable images must be selected and their facts and associations combined into questions that require cross-image comparison.

\begin{figure}[!t]
    \centering
    \includegraphics[width=\linewidth]{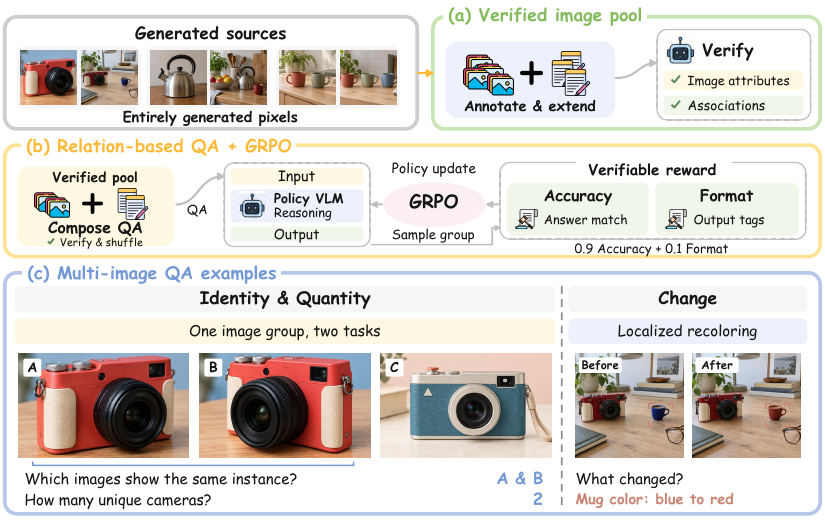}
    \caption{Overview of ReSynth.
    (a) Verified image facts and associations form a reusable pool.
    (b) Selected images yield checked QA for GRPO training.
    (c) Example identity, counting, and change tasks.}
    \label{fig:method_overview}
\end{figure}

To address these challenges, we introduce ReSynth (Relation-based Synthesis), which constructs multi-image supervision from verified visual relations in generated images.
\rev{It combines Qwen3-VL-8B's captioning, grounding, and counting capabilities with Qwen-Image-2.1's generation and editing capabilities.}
Its three-stage framework builds a reusable image pool and composes QA from it (Figure~\ref{fig:method_overview}).
First, ReSynth samples compatible combinations of object categories, scenes, activities, and physical relations before an LLM expands them into captions for source-image generation.
This controls source-image content while allowing the same category to appear in different settings for later comparison.
\rev{It then records what each image actually shows through single-image fact verification, rather than treating the plans as labels.}
\rev{Second, ReSynth extends accepted sources through generation and editing, complemented by geometric and temporal construction, and retains derived images and associations after branch-specific checks.}
Finally, ReSynth \rev{combines verified single-image facts with these constructed associations to compose multi-image QA}.
Task-specific templates and computations derive answers from the accepted records, including visible counts, instance associations, geometric projections, and temporal order.

Using this pipeline, we construct \rev{440K} multi-image examples across \rev{seven} core abilities entirely from generated images.
Although the ReSynth training set contains no real images, training Qwen2.5-VL-7B~\citep{bai2025qwen25vl} on it with GRPO~\citep{shao2024deepseekmath,meng2025mmeureka} improves the model's average score across BLINK, MUIRBench, and VLM2-Bench by \rev{17.73} points, with a \rev{6.10}-point gain on Qwen3-VL-8B~\citep{bai2025qwen3vl}.
\rev{Notably, the same 8B model used for perception during data construction benefits from the resulting supervision, outperforming Qwen3-VL-235B on BLINK and MUIRBench.}

Our contributions are:\par
\begingroup
\setlength{\leftmargini}{1.5em}
\setlength{\topsep}{0pt}
\setlength{\partopsep}{0pt}
\begin{enumerate}
\setlength{\labelwidth}{1em}
\setlength{\labelsep}{0.5em}
\renewcommand{\makelabel}[1]{#1\hfil}
\renewcommand{\labelenumi}{\textbf{\arabic{enumi}.}}
\setlength{\itemsep}{3pt}
\setlength{\parskip}{0pt}
\setlength{\parsep}{0pt}
\item We introduce ReSynth, a relation-based synthesis framework that \rev{combines single-image perception with controllable image generation to construct multi-image supervision from verified facts and constructed associations}.
Its reusable image pool allows the same visual evidence to support different questions.
\item We construct a \textbf{\rev{440K}}-example training corpus entirely from generated images, covering identity, change, quantity, composition, correspondence\rev{, viewpoint, and temporal relations}.
\item We show that reinforcement learning on generated images alone can improve multi-image understanding.
ReSynth raises the average score across BLINK, MUIRBench, and VLM2-Bench by \textbf{\rev{17.73}} points on Qwen2.5-VL-7B and \textbf{\rev{6.10}} points on Qwen3-VL-8B.
\end{enumerate}
\endgroup

\section{Related Work}

\paragraph{Multi-image evaluation.}
Multi-image benchmarks test whether models can connect evidence across images.
BLINK~\citep{fu2024blink} casts 14 perception tasks, including relative depth and visual correspondence, as multiple-choice questions.
MUIRBench~\citep{wang2024muirbench} covers ten multi-image relation categories with paired answerable and unanswerable examples.
VLM2-Bench~\citep{zhang2025vlm2bench} focuses on linking people and objects across views, while MMIU~\citep{meng2024mmiu} spans 52 tasks over seven relation types.

\paragraph{Existing multi-image supervision.}
Existing methods primarily construct supervision from available images.
Mantis~\citep{jiang2024mantis} aggregates interleaved instructions, SMiR~\citep{li2025smir} retrieves correlated images, and LLaVA-NeXT-Interleave~\citep{li2024llavainterleave} combines image, video, and 3D inputs.
PeRL~\citep{perl2025} trains with image permutations, MiCo~\citep{mico2025} contrasts image groups, and CGC~\citep{cgc2026} generates grounded examples from detections.
Multi-SpatialMLLM~\citep{xu2025multispatial} uses annotated 3D and 4D scenes, while ReMoT~\citep{wan2026remot} derives motion triplets from video.
Their available images and annotations constrain which identities, states, and viewpoints can be related.

\paragraph{Synthetic data for LLMs and VLMs.}
Self-Instruct~\citep{wang2023selfinstruct}, STaR~\citep{zelikman2022star}, and Magpie~\citep{xu2025magpie} generate training data for language models.
Visual generation supports synthetic scenes~\citep{team2025zimage}, local edits~\citep{tencenthunyuanfoundationmodelteam2026hunyuanimage30technicalreport}, and new viewpoints~\citep{wu2025qwenimage,odin2026multipleangles}.
R1-SyntheticVL~\citep{cads2026} studies generated images for multimodal learning, while VisionFoundry~\citep{zhou2026visionfoundry} verifies task-directed visual examples.
ReSynth verifies and reuses cross-image associations to construct multi-image QA.

\section{Method}

ReSynth \rev{turns single image perception and controllable image generation into cross image supervision: it} builds a reusable image pool with verified facts and checked \rev{cross image} associations, then composes multi-image QA from it (Figure~\ref{fig:method_overview}).

\begingroup
\setlength{\leftmargini}{1.5em}
\begin{enumerate}
    \setlength{\labelwidth}{1em}
    \setlength{\labelsep}{0.5em}
    \renewcommand{\makelabel}[1]{#1\hfil}
    \renewcommand{\labelenumi}{\textbf{\arabic{enumi}.}}
    \setlength{\itemsep}{4pt}
    \setlength{\parskip}{0pt}
    \setlength{\parsep}{0pt}

\item \textbf{Structured Source Generation \rev{and Fact Verification}.}
Compatible scene elements are combined into plans that specify the intended content and candidate contrasts between sources.
A language model expands each plan into a caption, which an image generation model renders as a source image.
\rev{Single image perception then checks the visible objects, locations, counts, and attributes, retaining usable sources with their observed facts.}

\item \textbf{\rev{Relation Construction through Generation and Editing}.}
\rev{Accepted sources are extended through generation and editing, complemented by geometric and temporal construction.}
\rev{Branch specific checks retain usable derived images and their change, instance, correspondence, viewpoint, or temporal associations.}

\item \textbf{Composing Multi-Image Supervision.}
Each task type selects and combines images using their verified \rev{single image} facts and checked associations to construct questions and answers.
\end{enumerate}
\endgroup

\subsection{\texorpdfstring{Structured Source Generation \rev{and Fact Verification}}{Structured Source Generation and Fact Verification}}
\label{sec:source_pool}

\begin{figure}[!t]
    \centering
    \includegraphics[width=0.95\linewidth]{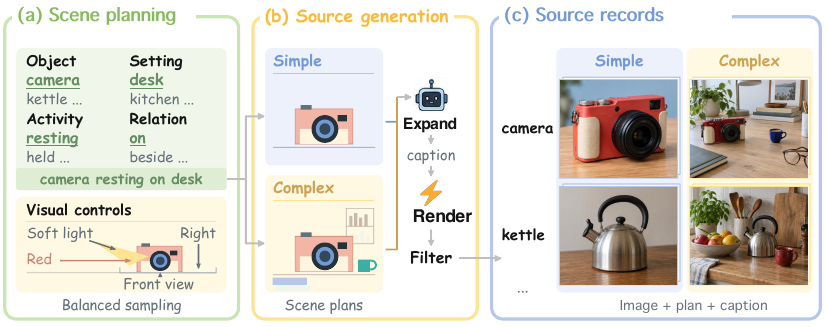}
    \caption{Structured source generation. (a)~Compatible elements form scene plans. (b)~Simple and complex plans become captions and images. (c)~Accepted sources retain their plans and captions.}
    \label{fig:source_generation}
\end{figure}

\paragraph{Scene planning.}
To produce sources that can be reused across questions, ReSynth plans their content before generating images.
It samples a main object category, a compatible setting and activity, and a visible physical relation from predefined pools; appearance, lighting, viewpoint, and placement provide further variation.
Each plan records intended attributes that can be shared or contrasted across source images.
For example, two plans can share the bicycle category while placing one bicycle on a street and another beside a workbench in a repair shop.
\rev{Sampling favors less used compatible combinations to broaden coverage of settings and relations for each category.}
Figure~\ref{fig:source_generation} shows how these plans are expanded into captions, rendered as images, and filtered for the source pool.

\paragraph{Scene richness control.}
Simple plans keep the main subject, activity, and essential context; complex plans add compatible objects across the scene.
A bicycle repair plan, for example, may contain only a bicycle, mechanic, and tools, or also tire racks, other bicycles, and participants.
Simple plans provide clear subjects for viewpoint generation, while complex plans provide more objects for annotation and comparison.

\paragraph{Source generation and records.}
\rev{Qwen3-VL-8B} expands each scene plan into a caption, which \rev{Qwen-Image-2.1} renders into a source image.
\rev{Each candidate image is stored with its plan and caption}, which describe the intended objects, setting, and relations.
The plans help retrieve candidate sources with shared or contrasting content; \rev{single image checks establish what the images actually show before these differences are used for QA}.

\paragraph{Source-image fact verification.}
Scene plans specify intended content, not image labels.
\rev{Qwen3-VL-8B} annotates object categories, boxes, visible counts, and attributes, \rev{using eight annotation samples and consistency checks to retain reliable facts}.
Counting samples whose repeated judgments do not reach agreement are discarded rather than retained with corrected labels.
These records guide later image selection and answer construction, while object boxes also locate edit targets and regions referenced by questions.
\rev{Images that pass the applicable source image checks are retained with their verified facts for subsequent relation and QA construction.}

\FloatBarrier
\subsection{\texorpdfstring{\rev{Relation Construction through Generation and Editing}}{Relation Construction through Generation and Editing}}
\label{sec:relation_construction}

ReSynth extends accepted sources using their verified facts to guide target selection.
\paragraph{Locally edited images.}
Object boxes guide target selection for local editing, including both the main subject and other visible objects, with area-based sampling to increase coverage of small targets.
An image editing model applies operations such as \texttt{add}, \texttt{remove}, \texttt{replace}, or \texttt{recolor} within the target region.
When appropriate, the region is cropped for editing and composited back into the source image to limit the affected area.
The source and edited images are compared to confirm the target change and surrounding consistency; accepted pairs retain the target location and observed before-and-after states.

\paragraph{\rev{Cross scene instance relations.}}
Sources with a complete, clearly visible main subject are selected to construct instance relations.
\rev{Qwen-Image-2.1 places the subject into new backgrounds through subject preserving editing or controlled redrawing, and creates look alike distractors with subtle detail changes.}
Visual checks assess subject identity and structural quality; accepted views are linked to the same \rev{or distinct} instance records for later QA.

\paragraph{\rev{Correspondence, viewpoint, and temporal relations.}}
\rev{Single image part grounding and geometric projections provide correspondence and viewpoint records, while generated videos and programmatic state sequences provide temporal order.}
Construction metadata determines the corresponding labels, with visual and geometric checks retaining usable examples.

\subsection{Composing Multi-Image Supervision}
\label{sec:compile_supervision}

The image pool does not bind each source or derived image to one question.
For each task type, ReSynth selects images from the pool, combining independent sources using their verified \rev{single image} facts or using images linked by \rev{checked construction associations}.
Scene plans help find candidate independent sources, but their observed content determines which comparisons are valid.
The resulting QA covers identity, change, quantity, composition, correspondence, viewpoint, and temporal relations, with the dataset distribution summarized in Table~\ref{tab:data_composition}.
These are operational groups for organizing construction, rather than mutually exclusive or exhaustive categories of multi-image reasoning.
Questions use multiple-choice, true/false, or integer-answer formats.

\paragraph{Image selection by task type.}
Identity and change tasks use existing associations: views of the same instance support matching and unique-instance counting, while before-and-after edit pairs support change questions.
Quantity and correspondence tasks can also combine independent sources according to their verified objects, counts, attributes, and relations.
For example, the bicycle plans in Section~\ref{sec:source_pool} suggest a scene-based comparison.
If the images confirm a bicycle beside a workbench in one source and a bicycle on a street in the other, the pair can support a question asking which image shows the bicycle beside the workbench.
Likewise, the same sources containing two and three cups can support either a total-count question with answer five or a comparison that selects the second image, without generating new images.
For composition tasks, ReSynth masks a source-image region and combines the original patch with distractor patches as answer candidates.

\paragraph{QA construction.}
When answers follow from verified records, as in count aggregation, templates and computations produce the QA.
A multimodal model supplies observations, part names, and candidate change descriptions, while task-specific builders arrange questions and options and derive reference answers from accepted records.
Geometric tasks use projection or camera-motion metadata, and temporal tasks use frame times or programmatic state order.
The same images can support different questions: an edited pair can be used to select the correct change description or judge whether a specified change occurred.
Distractors preserve some similarities but fail to satisfy the target relation, such as a different instance of the same category or a change description with the before and after states reversed.
\rev{Change tasks pair forward and reverse transitions with positive and negative judgments; task specific builders also control answer availability and candidate structure to reduce shortcuts. The recipe comparisons in Section~\ref{sec:qa_construction_analysis} examine these design choices.}

\paragraph{Relation and answer checks.}
Facts and associations must pass their corresponding checks before they are used for QA.
QA builders skip candidates with missing records, insufficient distractors, invalid geometry or order, or inconsistent answer mappings.

\begin{table}[!t]
\centering
\caption{Distribution of approximately \rev{440K} ReSynth examples across \rev{seven} abilities.}
\label{tab:data_composition}
\RSTTableSetup
\renewcommand{\arraystretch}{1.26}
\setlength{\tabcolsep}{5pt}
\begin{tabularx}{\textwidth}{@{} l >{\RSTNumberFont}c >{\raggedright\arraybackslash}X >{\RSTNumberFont}r @{}}
\toprule
\textbf{Ability} & \textbf{\# Ops.} & \textbf{QA operations} & \textbf{Share} \\
\midrule
Identity & \rev{4} & Same-object/person judgment; instance grouping; unique counting & \rev{8\%} \\
Change & 3 & Change matching; change tracking; spot-the-difference & \rev{17\%} \\
Quantity & \rev{1} & Cross-image count aggregation and comparison & \rev{8\%} \\
Composition & 1 & Missing-patch recovery & \rev{9\%} \\
Correspondence & \rev{3} & \rev{Point, semantic-part, and functional} matching & \rev{29\%} \\
\rev{Viewpoint} & \rev{2} & \rev{Camera turn; surround views} & \rev{19\%} \\
\rev{Temporal} & \rev{1} & \rev{Temporal action/state ordering} & \rev{10\%} \\
\midrule
\textbf{Total} & \textbf{\rev{15}} & & \textbf{100\%} \\
\bottomrule
\end{tabularx}
\end{table}

\subsection{Training}
We train the model with GRPO~\citep{shao2024deepseekmath}, using rule-based correctness and format rewards as in DeepSeek-R1~\citep{deepseek2025r1}.
For each question, the model samples a group of responses.
Each response places visual reasoning inside \texttt{<think>...</think>}, followed by the final answer inside \texttt{<answer>...</answer>}.
Given a response $o$ and reference answer $a$, the reward is
\begin{equation}
r(o,a)=0.9\,r_{\mathrm{acc}}(o,a)+0.1\,r_{\mathrm{fmt}}(o),
\label{eq:reward}
\end{equation}
where $r_{\mathrm{acc}}$ scores the correctness of an option letter, T/F label, or integer, and $r_{\mathrm{fmt}}$ checks the required output format.
The reasoning content is not scored separately.
We use the same objective across all task types.

\section{Experiments}

We test whether entirely generated images and their verified cross-image relations can provide effective training supervision for multi-image understanding.
We first evaluate ReSynth across models and tasks, then examine training scale and the construction of multi-image QA from established relations.
Our experiments address three questions:
\begingroup
\setlength{\leftmargini}{1.5em}
\begin{enumerate}
    \setlength{\labelwidth}{1em}
    \setlength{\labelsep}{0.5em}
    \renewcommand{\makelabel}[1]{#1\hfil}
    \renewcommand{\labelenumi}{\textbf{\arabic{enumi}.}}
    \setlength{\itemsep}{3pt}
    \setlength{\parskip}{0pt}
    \setlength{\parsep}{0pt}
    \item \textbf{Effectiveness and transfer:} Can supervision constructed entirely from generated images improve multi-image understanding across VLMs and benefit different tasks?
    \item \textbf{\rev{Training scale and relation-based QA construction:}} \rev{How does performance evolve over training, and how can established relations be used to construct effective multi-image QA?}
    \item \textbf{Discussion:} \rev{How do data diversity and training strategy affect the benefits of synthetic multi-image supervision?}
\end{enumerate}
\endgroup

\subsection{Experimental Setup}

\paragraph{Training Data.}
ReSynth contains \rev{440K} examples covering \rev{seven} multi-image abilities, with the data composition summarized in Table~\ref{tab:data_composition}.
\rev{Qwen3-VL-8B performs caption expansion and single-image fact annotation, while Qwen-Image-2.1 generates and edits source images to construct diverse scenes, changes, and instance relations.}
\rev{In addition, geometric projections provide correspondence and viewpoint supervision; videos generated by Wan-series models are sampled into frames for temporal-order supervision; and programmatically rendered pages, charts, and controlled state sequences supplement change recognition and order understanding.}

\paragraph{Models and Training.}
We report experiments on Qwen2.5-VL-7B-Instruct~\citep{bai2025qwen25vl} and Qwen3-VL-8B-Instruct~\citep{bai2025qwen3vl} to evaluate ReSynth on two generations of Qwen-VL models.
All models undergo full-parameter fine-tuning with the same GRPO training settings and a learning rate of $10^{-6}$.
We select checkpoints based on training reward: Qwen3-VL-8B's reward plateaus around the midpoint, while Qwen2.5-VL-7B's continues to increase, so we use the half-scale and full-scale checkpoints, respectively.

\Needspace{8\baselineskip}
\paragraph{Evaluation Protocol.}
We use VLMEvalKit~\citep{duan2024vlmevalkit} to evaluate on BLINK~\citep{fu2024blink}, MUIRBench~\citep{wang2024muirbench}, and the eight image-based tasks of \mbox{VLM2-Bench~\citep{zhang2025vlm2bench}}.
For BLINK, we report overall, single-image, and multi-image results on 1,901, 793, and 1,108 examples, respectively.
We report overall accuracy for MUIRBench and the macro-average over the eight VLM2-Bench tasks using their official scoring rules.
Results for prior methods are taken from their papers and marked accordingly in the table.

\subsection{Main Results}

\paragraph{Overall Performance.}
\begin{table}[!t]
  \centering
  \begin{threeparttable}
    \caption{Results on BLINK \RSTciteBlink, MUIRBench \RSTciteMuir,
    and VLM2-Bench \RSTciteVlm. Avg. (Average) is the mean of BLINK,
    MUIRBench, and VLM2. Scores are accuracies (\%).}
    \label{tab:main}
    \RSTTableSetup
    \newcommand{\resultgain}[1]{\makebox[2.2em][l]{\,{\scriptsize\textcolor{red!85!black}{\textbf{$\uparrow$#1}}}}}
    \newcommand{\nogain}{\makebox[2.2em][l]{}}
    \setlength{\tabcolsep}{6pt}
    \renewcommand{\arraystretch}{1.12}
    \begin{tabularx}{\linewidth}{@{}>{\raggedright\arraybackslash}X>{\RSTNumberFont}r*{4}{>{\RSTNumberFont}r}@{}}
      \toprule
      \textbf{Model / Method} & \textbf{\#Train}
      & \textbf{BLINK}\nogain & \textbf{MUIR}\nogain & \textbf{VLM2}\nogain & \textbf{Avg.}\nogain \\
      \midrule
      GPT-4o\textsuperscript{\textdagger}~\citep{openai2024gpt4o} & \RSTna
        & 68.0\nogain & 68.0\nogain & 59.6\nogain & 65.2\nogain \\
      Qwen2.5-VL-7B \RSTciteQwen & \RSTna
        & \rev{53.3}\nogain & \rev{56.3}\nogain & \rev{52.0}\nogain & \rev{53.9}\nogain \\
      \rev{Qwen3-VL-8B~\citep{bai2025qwen3vl}} & \RSTna
        & \rev{65.5}\nogain & \rev{73.7}\nogain & \rev{78.8}\nogain & \rev{72.6}\nogain \\
      \rev{Qwen3-VL-235B~\citep{bai2025qwen3vl}} & \RSTna
        & \rev{70.7}\nogain & \rev{72.8}\nogain & -\nogain & -\nogain \\
      MM-Eureka\textsuperscript{\textdagger} \RSTciteEureka & 15.6K
        & 54.4\nogain & 60.6\nogain & 59.0\nogain & 58.0\nogain \\
      PeRL\textsuperscript{\textdagger} \RSTcitePerl & 58K
        & 58.5\nogain & -\nogain & -\nogain & -\nogain \\
      MiCo\textsuperscript{\textdagger} \RSTciteMico & 87K
        & 57.2\nogain & 60.5\nogain & 63.2\nogain & 60.3\nogain \\
      CGC-7B\textsuperscript{\textdagger} \RSTciteCgc & 24K
        & 56.2\nogain & 60.9\nogain & 63.7\nogain & 60.3\nogain \\
      \midrule
      \rowcolor{RSTOurs}
      \textbf{ReSynth (Qwen2.5-VL-7B)} & \rev{440K}
        & \rev{68.6}\resultgain{15.3} & \rev{70.5}\resultgain{14.3} & \rev{75.6}\resultgain{23.6} & \rev{71.6}\resultgain{17.7} \\
      \rowcolor{RSTOurs}
      \rev{\textbf{ReSynth (Qwen3-VL-8B)}} & \rev{440K}
        & \rev{74.8}\resultgain{9.3} & \rev{80.1}\resultgain{6.4} & \rev{81.4}\resultgain{2.6} & \rev{78.7}\resultgain{6.1} \\
      \bottomrule
    \end{tabularx}
  \end{threeparttable}
\end{table}

As shown in Table~\ref{tab:main}, training Qwen2.5-VL-7B \rev{and Qwen3-VL-8B} with ReSynth improves performance on BLINK, MUIRBench, and VLM2-Bench \rev{(+17.7 and +6.1 average points, respectively)}.
\rev{Notably, ReSynth-trained Qwen3-VL-8B outperforms Qwen3-VL-235B by 4.1 points on BLINK and 7.3 points on MUIRBench.}
Its average across the three benchmarks also exceeds the reported scores of MiCo and CGC, which construct supervision from existing visual data, showing that generated images can provide effective supervision for multi-image understanding.
On BLINK, gains appear in both the multi-image and single-image portions, indicating that the benefit extends to single-image understanding.

\paragraph{Transfer to Other Tasks.}
%
\rev{Table~\ref{tab:single_and_transfer}} reports \rev{MMIU~\citep{meng2024mmiu}, DiCoBench,} MMStar~\citep{chen2024mmstar}, MMMU~\citep{yue2023mmmu}, MathVista~\citep{lu2024mathvista}, and HallusionBench~\citep{guan2024hallusionbench} \rev{across Qwen2.5-VL-7B and Qwen3-VL-8B} to assess \rev{transfer and} single-image performance.
ReSynth training transfers to additional multi-image benchmarks: Qwen3-VL-8B gains +5.1 on MMIU and +6.4 on DiCoBench, while Qwen2.5-VL-7B gains \rev{+4.39 on MMIU and} +8.0 on DiCoBench.
\rev{Notably, the DiCoBench gains are achieved without constructing training QA specifically for this benchmark, indicating cross-task transfer.}
Qwen3-VL-8B improves on all four single-image benchmarks; Qwen2.5-VL-7B improves on MMStar and MathVista, remains unchanged on HallusionBench, and decreases by 1.4 points on MMMU.

\begin{figure}[!t]
    \centering
    \includegraphics[width=\linewidth]{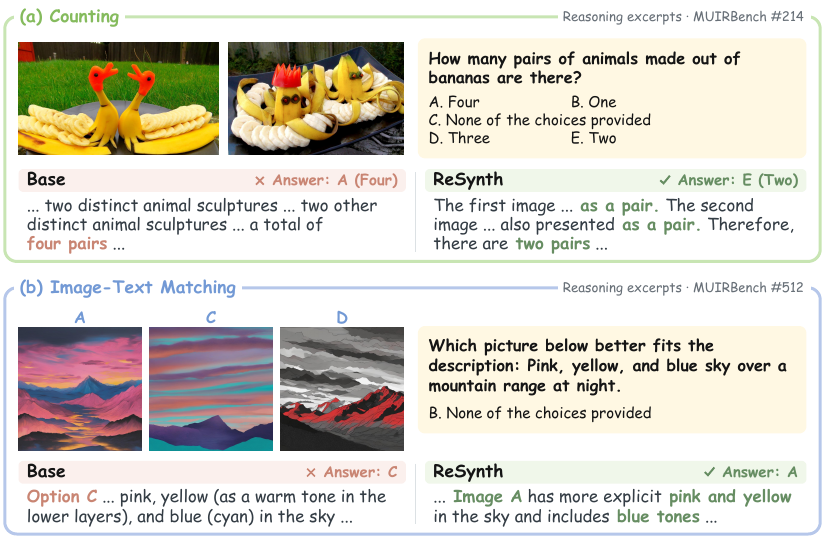}
\caption{Qualitative comparisons on MUIRBench with matched prompts and inference settings for base and ReSynth-trained Qwen2.5-VL-7B.
ReSynth (a) counts pairs across images and (b) matches the specified sky colors.
Images and questions are unmodified; both panels show reasoning excerpts, with omissions marked by ellipses.}
\label{fig:muir_qualitative_cases}
\end{figure}

\paragraph{Qualitative Comparison.}
Figure~\ref{fig:muir_qualitative_cases} compares responses from the base and ReSynth-trained models under the same inference settings.
On two MUIRBench examples, the trained model correctly counts sculpture pairs and matches images to the specified sky colors, correcting errors made by the base model.

\begin{table}[!t]
\centering
\caption{\rev{Additional multi-image and} single-image benchmark results for Qwen2.5-VL-7B \rev{and Qwen3-VL-8B}.
Scores are percentages; $\Delta$ denotes ReSynth minus Base.}
\label{tab:single_and_transfer}
\RSTTableSetup
\renewcommand{\arraystretch}{1.15}
\setlength{\tabcolsep}{4pt}
\begin{tabularx}{\textwidth}{@{}>{\raggedright\arraybackslash}X *{6}{>{\RSTNumberFont\centering\arraybackslash}X}@{}}
\toprule
& \multicolumn{2}{c}{\rev{\textbf{Additional multi-image}}} & \multicolumn{4}{c}{\textbf{Single-image benchmarks}} \\
\cmidrule(lr){2-3}\cmidrule(l){4-7}
\textbf{Model} & \RSThead{\rev{MMIU}} & \RSThead{\rev{DiCoBench}} & \RSThead{MMStar} & \RSThead{MMMU} & \RSThead{MathVista} & \RSThead{Hallusion} \\
\midrule
\multicolumn{7}{@{}l}{\textbf{Qwen2.5-VL-7B}} \\
Base & \rev{52.94} & \rev{35.4} & \rev{64.7} & \rev{53.4} & \rev{67.2} & \rev{55.6} \\
\rowcolor{RSTOurs}
\textbf{ReSynth} & \rev{57.33} & \rev{43.4} & \rev{65.4} & \rev{52.0} & \rev{68.1} & \rev{55.6} \\
$\Delta$ & \gain{+4.39} & \gain{+8.0} & \gain{+0.7} & \rev{$-1.4$} & \gain{+0.9} & \rev{+0.0} \\
\midrule
\multicolumn{7}{@{}l}{\rev{\textbf{Qwen3-VL-8B}}} \\
Base & \rev{57.8} & \rev{40.8} & \rev{70.9} & \rev{62.2} & \rev{75.5} & \rev{57.8} \\
\rowcolor{RSTOurs}
\textbf{ReSynth} & \rev{62.9} & \rev{47.2} & \rev{72.3} & \rev{66.4} & \rev{76.9} & \rev{61.4} \\
$\Delta$ & \gain{+5.1} & \gain{+6.4} & \gain{+1.4} & \gain{+4.2} & \gain{+1.4} & \gain{+3.6} \\
\bottomrule
\end{tabularx}
\end{table}

\subsection{Data Scale and \rev{QA Construction}}

\paragraph{Effect of Data Scale.}
Figure~\ref{fig:data_analysis} tracks performance across fractions of the \rev{440K} corpus.
Performance rises sharply by $1/6$ scale, \rev{while the optimal scale depends on backbone capability: Qwen2.5-VL-7B benefits from the full corpus and peaks at $1$, whereas the stronger Qwen3-VL-8B absorbs the relational supervision earlier and peaks at $1/2$}.

\paragraph{\rev{Effect of Relation-Based QA Construction.}}
\label{sec:qa_construction_analysis}
\rev{Known visual relations do not guarantee effective QA. Models may exploit keywords or candidate-answer cues. We compare QA construction recipes using approximately 10K examples each. For change tasks, we add reverse-direction descriptions as negatives. This requires distinguishing the direction of a change. The revised recipe raises change-tracking accuracy from 44.75 to 72.15. For patch completion, we stop using visible source regions as distractors. This removes a cue for rejecting candidates without matching the missing patch. The revised recipe raises accuracy from 40.00 to 84.67. These recipe-level gains highlight the importance of question and candidate design when turning established relations into supervision.}

\enlargethispage{1.5\baselineskip}
\subsection{Discussion}
\label{sec:discussion}

\paragraph{Subject Diversity in Captions.}
\label{sec:caption_diversity}
To test whether prompting alone broadens the range of visual subjects, we use Qwen3.6-27B to generate 500 image descriptions for each of three prompts under identical sampling settings:
\begingroup
\setlength{\leftmargini}{1.9em}
\begin{enumerate}
    \renewcommand{\labelenumi}{(\arabic{enumi})}
    \setlength{\labelwidth}{1.4em}
    \setlength{\labelsep}{0.5em}
    \renewcommand{\makelabel}[1]{#1\hfil}
    \setlength{\topsep}{0pt}
    \setlength{\partopsep}{0pt}
    \setlength{\itemsep}{3pt}
    \setlength{\parskip}{0pt}
    \setlength{\parsep}{0pt}
\item Free caption: choose any subject and scene for a realistic photograph.
\item Distinctive subject: choose a visually distinctive subject from any domain and describe its appearance and setting.
\item Plan before captioning: consider subjects from several unrelated domains, then choose a less common, visually concrete subject to describe.
\end{enumerate}
\endgroup
\Needspace{6\baselineskip}
For each prompt, we measure the share of descriptions belonging to its five most frequent subject categories; a higher share means greater category concentration.
The free-caption prompt yields 29.4\%, while asking for a distinctive subject raises this share to 37.8\%, with ``tree'' alone accounting for 16.4\%.
Planning before captioning lowers the top-five share to 11.2\%, but the descriptions still converge in appearance: 48\% depict rusted or corroded objects, and 32\% use close-up framing.
Thus, prompting can reduce repetition of subject categories while leaving a recurring visual style.
ReSynth instead samples object categories and compatible scenes before caption expansion to guide coverage of both subjects and settings (Section~\ref{sec:source_pool}).

\paragraph{Matching Data Difficulty to Model Capability.}
\begin{figure}[!t]
\centering
\includegraphics[width=\linewidth]{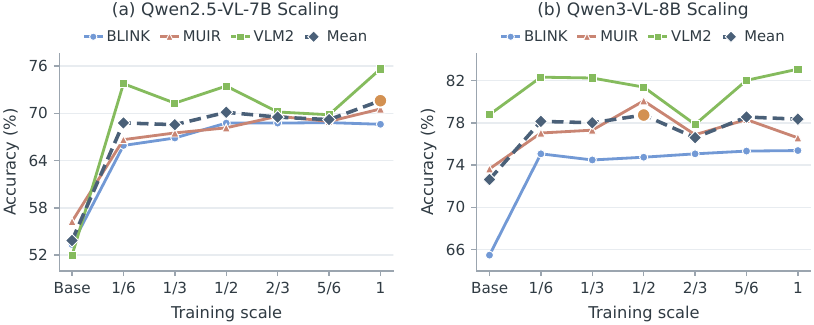}
\caption{Effect of training data scale on two model backbones across fractions of the \rev{440K} ReSynth corpus.
(a) Scores of Qwen2.5-VL-7B on BLINK, MUIRBench, VLM2-Bench, and their mean \rev{(peaking at full scale, $1$)}.
(b) Scores of Qwen3-VL-8B on the same three benchmarks and their mean \rev{(peaking at half scale, $1/2$)}.}
\label{fig:data_analysis}
\end{figure}

The ReSynth corpus improves Qwen2.5-VL-7B \rev{by +17.7 points}, while Qwen3-VL-8B, which already has stronger cross-image reasoning ability, \rev{gains +6.1 points and reaches its peak at half the training scale ($1/2$ vs.\ $1$ in Figure~\ref{fig:data_analysis})}.
\rev{This earlier saturation may reflect Qwen3-VL-8B's stronger initial capabilities, which leave less room for improvement on the current data.}

\paragraph{\rev{Bootstrapping Multi-Image Reasoning from Single-Image Perception.}}
\rev{Although Qwen3-VL-8B serves as the perception model during synthesis, training on ReSynth still improves Qwen3-VL-8B by +6.1 points.
This is consistent with task asymmetry: synthesis uses single-image observations, local checks, and known construction metadata, whereas inference requires recovering relations across shuffled image groups without that metadata.}

\begin{table}[!t]
\centering
\caption{\rev{Comparison of supervised fine-tuning (SFT) and reinforcement learning (RL) on the ReSynth corpus.
\textsuperscript{\textdagger}SFT overfits to training option templates and fails to follow VLM2-Bench output formats.}}
\label{tab:sft_vs_rl}
\RSTTableSetup
\renewcommand{\arraystretch}{1.15}
\setlength{\tabcolsep}{5pt}
\begin{tabularx}{\textwidth}{@{}>{\raggedright\arraybackslash}X *{6}{>{\RSTNumberFont\centering\arraybackslash}X}@{}}
\toprule
& \multicolumn{3}{c}{\textbf{Qwen2.5-VL-7B}} & \multicolumn{3}{c}{\rev{\textbf{Qwen3-VL-8B}}} \\
\cmidrule(lr){2-4}\cmidrule(l){5-7}
\textbf{Method} & \RSThead{BLINK} & \RSThead{MUIR} & \RSThead{VLM2} & \RSThead{BLINK} & \RSThead{MUIR} & \RSThead{VLM2} \\
\midrule
Base & \rev{53.3} & \rev{56.3} & \rev{52.0} & \rev{65.5} & \rev{73.7} & \rev{78.8} \\
SFT & \rev{67.1} & \rev{58.0} & \rev{0.0\textsuperscript{\textdagger}} & \rev{62.1} & \rev{58.4} & \rev{0.0\textsuperscript{\textdagger}} \\
\rowcolor{RSTOurs}
\textbf{\rev{RL}} & \rev{\textbf{68.6}} & \rev{\textbf{70.5}} & \rev{\textbf{75.6}} & \rev{\textbf{74.8}} & \rev{\textbf{80.1}} & \rev{\textbf{81.4}} \\
\bottomrule
\end{tabularx}
\end{table}

\paragraph{\rev{Reinforcement Learning vs.\ Supervised Fine-Tuning.}}
\rev{As shown in Table~\ref{tab:sft_vs_rl}, SFT approaches RL on BLINK for Qwen2.5-VL-7B but falls behind on MUIRBench and VLM2-Bench, degrades Qwen3-VL-8B below Base, and fails to follow VLM2-Bench output formats (0.0\%\textsuperscript{\textdagger}).
By rewarding verified answers while letting the model sample its own comparison chains, RL learns transferable multi-image reasoning without format overfitting.}

\section{Conclusion}
ReSynth demonstrates that entirely generated images can support multi-image understanding.
Its relation-based pipeline generates source images, verifies their visible content, and checks the changes and same-instance links created through editing and viewpoint generation.
It retains these facts and links in a reusable pool, allowing the same images to support different QA tasks across \rev{seven} abilities.
GRPO training on the resulting corpus improves Qwen2.5-VL-7B on BLINK, MUIRBench, and VLM2-Bench, with an average above the results reported by MiCo and CGC.
This shows that generated images can provide useful training supervision without real images in the ReSynth training set.

\rev{The recipe comparisons show that organizing established relations into questions and candidate answers affects the effectiveness of supervision.}
Stronger models gain less from additional examples, and viewpoint generation remains difficult in cluttered scenes.
Future work will construct harder, more diverse cross-image relations and improve consistency across generated views.

%

\section*{AI use statement}

Generative AI is part of our data-construction pipeline.
We use language and multimodal models to expand scene plans into captions, annotate visible content, and help generate and check candidate QA.
Image generation and editing models create source images, local changes, and alternative views.
Generated outputs undergo the visual verification and QA checks described in this paper, and training answers are determined from verified image facts and retained construction records.
We also use language models to improve manuscript clarity and assist with code development.
The authors reviewed AI-assisted text and code and take responsibility for the reported results and final content of this work.

\section*{Ethics statement}

This work does not involve human-subject studies or the direct collection of personal photographs.
The training corpus consists of generated images, while existing public benchmarks are used only for evaluation.
We will document the generation and editing provenance of the released corpus and its intended research use.

\section*{Reproducibility statement}

The Method and Experimental Setup sections describe the data construction, training, and evaluation procedures.
We plan to release the corresponding code to support reproducibility.

\bibliography{references}

@inproceedings{fu2024blink,
  title     = {{BLINK}: Multimodal Large Language Models Can See but Not Perceive},
  author    = {Fu, Xingyu and Hu, Yushi and Li, Bangzheng and Feng, Yu and Wang, Haoyu and Lin, Xudong and Roth, Dan and Smith, Noah A. and Ma, Wei-Chiu and Krishna, Ranjay},
  booktitle = {European Conference on Computer Vision (ECCV)},
  year      = {2024}
}

@inproceedings{wang2023selfinstruct,
  title     = {Self-Instruct: Aligning Language Models with Self-Generated Instructions},
  author    = {Wang, Yizhong and Kordi, Yeganeh and Mishra, Swaroop and Liu, Alisa and Smith, Noah A. and Khashabi, Daniel and Hajishirzi, Hannaneh},
  booktitle = {Proceedings of the 61st Annual Meeting of the Association for Computational Linguistics (ACL)},
  pages     = {13484--13508},
  year      = {2023}
}

@inproceedings{zelikman2022star,
  title     = {{STaR}: Bootstrapping Reasoning with Reasoning},
  author    = {Zelikman, Eric and Wu, Yuhuai and Mu, Jesse and Goodman, Noah D.},
  booktitle = {Advances in Neural Information Processing Systems (NeurIPS)},
  pages     = {15476--15488},
  year      = {2022}
}

@inproceedings{xu2025magpie,
  title     = {{Magpie}: Alignment Data Synthesis from Scratch by Prompting Aligned {LLMs} with Nothing},
  author    = {Xu, Zhangchen and Jiang, Fengqing and Niu, Luyao and Deng, Yuntian and Poovendran, Radha and Choi, Yejin and Lin, Bill Yuchen},
  booktitle = {International Conference on Learning Representations (ICLR)},
  year      = {2025}
}

@article{wang2024muirbench,
  title   = {{MuirBench}: A Comprehensive Benchmark for Robust Multi-image Understanding},
  author  = {Wang, Fei and Fu, Xingyu and Huang, James Y. and Li, Zekun and Liu, Qin and Liu, Xiaogeng and Ma, Mingyu Derek and Xu, Nan and Zhou, Wenxuan and Zhang, Kai and Yan, Tianyi Lorena and Mo, Wenjie Jacky and Liu, Hsiang-Hui and Lu, Pan and Li, Chunyuan and Xiao, Chaowei and Chang, Kai-Wei and Roth, Dan and Zhang, Sheng and Poon, Hoifung and Chen, Muhao},
  journal = {arXiv preprint arXiv:2406.09411},
  year    = {2024}
}

@article{zhang2025vlm2bench,
  title   = {{VLM2-Bench}: A Closer Look at How Well {VLMs} Implicitly Link Explicit Matching Visual Cues},
  author  = {Zhang, Jianshu and Yao, Dongyu and Pi, Renjie and Liang, Paul Pu and Fung, Yi R.},
  journal = {arXiv preprint arXiv:2502.12084},
  year    = {2025}
}

@inproceedings{mico2025,
  title     = {{MiCo}: Multi-image Contrast for Reinforcement Visual Reasoning},
  author    = {Chen, Xi and Zhu, Mingkang and Liu, Shaoteng and Wu, Xiaoyang and Xu, Xiaogang and Liu, Yu and Bai, Xiang and Zhao, Hengshuang},
  booktitle = {Advances in Neural Information Processing Systems (NeurIPS)},
  year      = {2025}
}

@inproceedings{perl2025,
  title     = {{PeRL}: Permutation-Enhanced Reinforcement Learning for Interleaved Vision-Language Reasoning},
  author    = {Zhang, Yizhen and Ding, Yang and Zhang, Shuoshuo and Zhang, Xinchen and Li, Haoling and Li, Zhong-zhi and Wang, Peijie and Wu, Jie and Ji, Lei and Shen, Yelong and Yang, Yujiu and Gong, Yeyun},
  booktitle = {Advances in Neural Information Processing Systems (NeurIPS)},
  year      = {2025}
}

@article{cads2026,
  title   = {{R1-SyntheticVL}: Is Synthetic Data from Generative Models Ready for Multimodal Large Language Model?},
  author  = {Zhang, Jingyi and Lin, Tianyi and Yao, Huanjin and Lan, Xiang and Liu, Shunyu and Huang, Jiaxing},
  journal = {arXiv preprint arXiv:2602.03300},
  year    = {2026}
}

@article{bai2025qwen25vl,
  title   = {{Qwen2.5-VL} Technical Report},
  author  = {Bai, Shuai and Chen, Keqin and Liu, Xuejing and Wang, Jialin and Ge, Wenbin and Song, Sibo and Dang, Kai and Wang, Peng and Wang, Shijie and Tang, Jun and Zhong, Humen and Zhu, Yuanzhi and Yang, Mingkun and Li, Zhaohai and Wan, Jianqiang and Wang, Pengfei and Ding, Wei and Fu, Zheren and Xu, Yiheng and Ye, Jiabo and Zhang, Xi and Xie, Tianbao and Cheng, Zesen and Zhang, Hang and Yang, Zhibo and Xu, Haiyang and Lin, Junyang},
  journal = {arXiv preprint arXiv:2502.13923},
  year    = {2025}
}

@article{shao2024deepseekmath,
  title   = {{DeepSeekMath}: Pushing the Limits of Mathematical Reasoning in Open Language Models},
  author  = {Shao, Zhihong and Wang, Peiyi and Zhu, Qihao and Xu, Runxin and Song, Junxiao and Bi, Xiao and Zhang, Haowei and Zhang, Mingchuan and Li, Y. K. and Wu, Y. and Guo, Daya},
  journal = {arXiv preprint arXiv:2402.03300},
  year    = {2024}
}

@article{meng2025mmeureka,
  title   = {{MM-Eureka}: Exploring the Frontiers of Multimodal Reasoning with Rule-based Reinforcement Learning},
  author  = {Meng, Fanqing and Du, Lingxiao and Liu, Zongkai and Zhou, Zhixiang and Lu, Quanfeng and Fu, Daocheng and Han, Tiancheng and Shi, Botian and Wang, Wenhai and He, Junjun and Zhang, Kaipeng and Luo, Ping and Qiao, Yu and Zhang, Qiaosheng and Shao, Wenqi},
  journal = {arXiv preprint arXiv:2503.07365},
  year    = {2025}
}

@article{jiang2024mantis,
  title   = {{Mantis}: Interleaved Multi-Image Instruction Tuning},
  author  = {Jiang, Dongfu and He, Xuan and Zeng, Huaye and Wei, Cong and Ku, Max and Liu, Qian and Chen, Wenhu},
  journal = {Transactions on Machine Learning Research},
  year    = {2024}
}

@article{zhou2026visionfoundry,
  title   = {{VisionFoundry}: Teaching {VLMs} Visual Perception with Synthetic Images},
  author  = {Zhou, Guanyu and Yin, Yida and Chai, Wenhao and Tong, Shengbang and Fu, Xingyu and Liu, Zhuang},
  journal = {arXiv preprint arXiv:2604.09531},
  year    = {2026}
}

@article{cgc2026,
  title   = {{CGC}: Compositional Grounded Contrast for Fine-Grained Multi-Image Understanding},
  author  = {Zheng, Lihao and Shao, Zhenwei and Zhou, Yu and Yang, Yan and Shen, Xintian and Chen, Jiawei and Ma, Hao and Wei, Tao},
  journal = {arXiv preprint arXiv:2604.22498},
  year    = {2026}
}

@article{meng2024mmiu,
  title   = {{MMIU}: Multimodal Multi-image Understanding for Evaluating Large Vision-Language Models},
  author  = {Meng, Fanqing and Wang, Jin and Li, Chuanhao and Lu, Quanfeng and Tian, Hao and Liao, Jiaqi and Zhu, Xizhou and Dai, Jifeng and Qiao, Yu and Luo, Ping and Zhang, Kaipeng and Shao, Wenqi},
  journal = {arXiv preprint arXiv:2408.02718},
  year    = {2024}
}

@article{yue2023mmmu,
  title   = {{MMMU}: A Massive Multi-discipline Multimodal Understanding and Reasoning Benchmark for Expert {AGI}},
  author  = {Yue, Xiang and Ni, Yuansheng and Zhang, Kai and Zheng, Tianyu and Liu, Ruoqi and Zhang, Ge and Stevens, Samuel and Jiang, Dongfu and Ren, Weiming and Sun, Yuxuan and others},
  journal = {arXiv preprint arXiv:2311.16502},
  year    = {2023}
}

@article{chen2024mmstar,
  title   = {Are We on the Right Way for Evaluating Large Vision-Language Models?},
  author  = {Chen, Lin and Li, Jinsong and Dong, Xiaoyi and Zhang, Pan and Zang, Yuhang and Chen, Zehui and Duan, Haodong and Wang, Jiaqi and Qiao, Yu and Lin, Dahua and Zhao, Feng},
  journal = {arXiv preprint arXiv:2403.20330},
  year    = {2024}
}

@inproceedings{lu2024mathvista,
  title     = {{MathVista}: Evaluating Mathematical Reasoning of Foundation Models in Visual Contexts},
  author    = {Lu, Pan and Bansal, Hritik and Xia, Tony and Liu, Jiacheng and Li, Chunyuan and Hajishirzi, Hannaneh and Cheng, Hao and Chang, Kai-Wei and Galley, Michel and Gao, Jianfeng},
  booktitle = {International Conference on Learning Representations (ICLR)},
  year      = {2024}
}

@inproceedings{guan2024hallusionbench,
  title     = {{HallusionBench}: An Advanced Diagnostic Suite for Entangled Language Hallucination and Visual Illusion in Large Vision-Language Models},
  author    = {Guan, Tianrui and Liu, Fuxiao and Wu, Xiyang and Xian, Ruiqi and Li, Zongxia and Liu, Xiaoyu and Wang, Xijun and Chen, Lichang and Huang, Furong and Yacoob, Yaser and Manocha, Dinesh and Zhou, Tianyi},
  booktitle = {Proceedings of the IEEE/CVF Conference on Computer Vision and Pattern Recognition (CVPR)},
  pages     = {14375--14385},
  year      = {2024}
}

@article{wu2025qwenimage,
  title   = {{Qwen-Image} Technical Report},
  author  = {Wu, Chenfei and Li, Jiahao and Zhou, Jingren and Lin, Junyang and Gao, Kaiyuan and Yan, Kun and Yin, Sheng-ming and Bai, Shuai and Xu, Xiao and Chen, Yilei and Chen, Yuxiang and Tang, Zecheng and Zhang, Zekai and Wang, Zhengyi and Yang, An and Yu, Bowen and Cheng, Chen and Liu, Dayiheng and Li, Deqing and Zhang, Hang and Meng, Hao and Wei, Hu and Ni, Jingyuan and Chen, Kai and Cao, Kuan and Peng, Liang and Qu, Lin and Wu, Minggang and Wang, Peng and Yu, Shuting and Wen, Tingkun and Feng, Wensen and Xu, Xiaoxiao and Wang, Yi and Zhang, Yichang and Zhu, Yongqiang and Wu, Yujia and Cai, Yuxuan and Liu, Zenan},
  journal = {arXiv preprint arXiv:2508.02324},
  year    = {2025}
}

@misc{odin2026multipleangles,
  title        = {{Qwen-Image-Edit-2511-Multiple-Angles-LoRA}},
  author       = {Odin, Lovis and fal},
  year         = {2026},
  howpublished = {Hugging Face model repository},
  url          = {https://huggingface.co/fal/Qwen-Image-Edit-2511-Multiple-Angles-LoRA}
}

@article{team2025zimage,
  title   = {{Z-Image}: An Efficient Image Generation Foundation Model with Single-Stream Diffusion Transformer},
  author  = {{Z-Image Team}},
  journal = {arXiv preprint arXiv:2511.22699},
  year    = {2025}
}

@misc{tencenthunyuanfoundationmodelteam2026hunyuanimage30technicalreport,
  title         = {{HunyuanImage 3.0 Technical Report}},
  author        = {{Tencent Hunyuan Foundation Model Team}},
  year          = {2026},
  eprint        = {2509.23951},
  archivePrefix = {arXiv},
  primaryClass  = {cs.CV},
  url           = {https://arxiv.org/abs/2509.23951}
}

@article{li2025smir,
  title   = {{SMIR}: Efficient Synthetic Data Pipeline To Improve Multi-Image Reasoning},
  author  = {Li, Andrew and Thapa, Rahul and Chalamala, Rahul and Wu, Qingyang and Chen, Kezhen and Zou, James},
  journal = {arXiv preprint arXiv:2501.03675},
  year    = {2025}
}

@article{openai2024gpt4o,
  title = {{GPT-4o System Card}},
  author = {{OpenAI}},
  journal = {arXiv preprint arXiv:2410.21276},
  year = {2024},
  url = {https://arxiv.org/abs/2410.21276}
}

@article{deepseek2025r1,
  title = {{DeepSeek-R1: Incentivizing Reasoning Capability in LLMs via Reinforcement Learning}},
  author = {{DeepSeek-AI} and Guo, Daya and Yang, Dejian and Zhang, Haowei and others},
  journal = {arXiv preprint arXiv:2501.12948},
  year = {2025},
  url = {https://arxiv.org/abs/2501.12948}
}

@article{bai2025qwen3vl,
  title = {{Qwen3-VL Technical Report}},
  author = {Bai, Shuai and Cai, Yuxuan and Chen, Ruizhe and Chen, Keqin and Chen, Xionghui and others},
  journal = {arXiv preprint arXiv:2511.21631},
  year = {2025},
  url = {https://arxiv.org/abs/2511.21631}
}

@article{duan2024vlmevalkit,
  title = {{VLMEvalKit: An Open-Source Toolkit for Evaluating Large Multi-Modality Models}},
  author = {Duan, Haodong and Fang, Xinyu and Yang, Junming and Zhao, Xiangyu and Ma, Zerun and others},
  journal = {arXiv preprint arXiv:2407.11691},
  year = {2024},
  url = {https://arxiv.org/abs/2407.11691}
}

@article{li2024llavainterleave,
  title = {{LLaVA-NeXT-Interleave: Tackling Multi-image, Video, and 3D in Large Multimodal Models}},
  author = {Li, Feng and Zhang, Renrui and Zhang, Hao and Zhang, Yuanhan and Li, Bo and others},
  journal = {arXiv preprint arXiv:2407.07895},
  year = {2024},
  url = {https://arxiv.org/abs/2407.07895}
}

@article{xu2025multispatial,
  title = {{Multi-SpatialMLLM: Multi-Frame Spatial Understanding with Multi-Modal Large Language Models}},
  author = {Xu, Runsen and Wang, Weiyao and Tang, Hao and Chen, Xingyu and Wang, Xiaodong and others},
  journal = {arXiv preprint arXiv:2505.17015},
  year = {2025},
  url = {https://arxiv.org/abs/2505.17015}
}

@article{wan2026remot,
  title = {{ReMoT: Reinforcement Learning with Motion Contrast Triplets}},
  author = {Wan, Cong and Guo, Zeyu and Li, Jiangyang and Dong, SongLin and Bai, Yifan and others},
  journal = {arXiv preprint arXiv:2603.00461},
  year = {2026},
  url = {https://arxiv.org/abs/2603.00461}
}
\bibliographystyle{iclr2027_conference}

\end{document}